%% file: main.tex
\documentclass[runningheads]{llncs}

\usepackage{cite}
\usepackage{amsmath,amssymb,amsfonts}
\usepackage{adjustbox, tabularx}
\usepackage{makecell, booktabs}
\usepackage{array, multirow}

\usepackage{algorithmic}
\usepackage{graphicx}
\graphicspath{{./Figs/}}
\usepackage{subcaption}
\usepackage{svg}
\usepackage{todonotes}
\usepackage{hyperref}
\usepackage{amsmath}
\usepackage{textcomp}
\usepackage{xcolor}
\usepackage{comment}
\usepackage{adjustbox}

\begin{document}

\title{Context matters for fairness - a case study on the effect of spatial distribution shifts}

\titlerunning{Context matters for fairness}
\author{Siamak Ghodsi\inst{1,2}\href{https://orcid.org/0000-0002-3306-4233}{\includegraphics[scale=0.009]{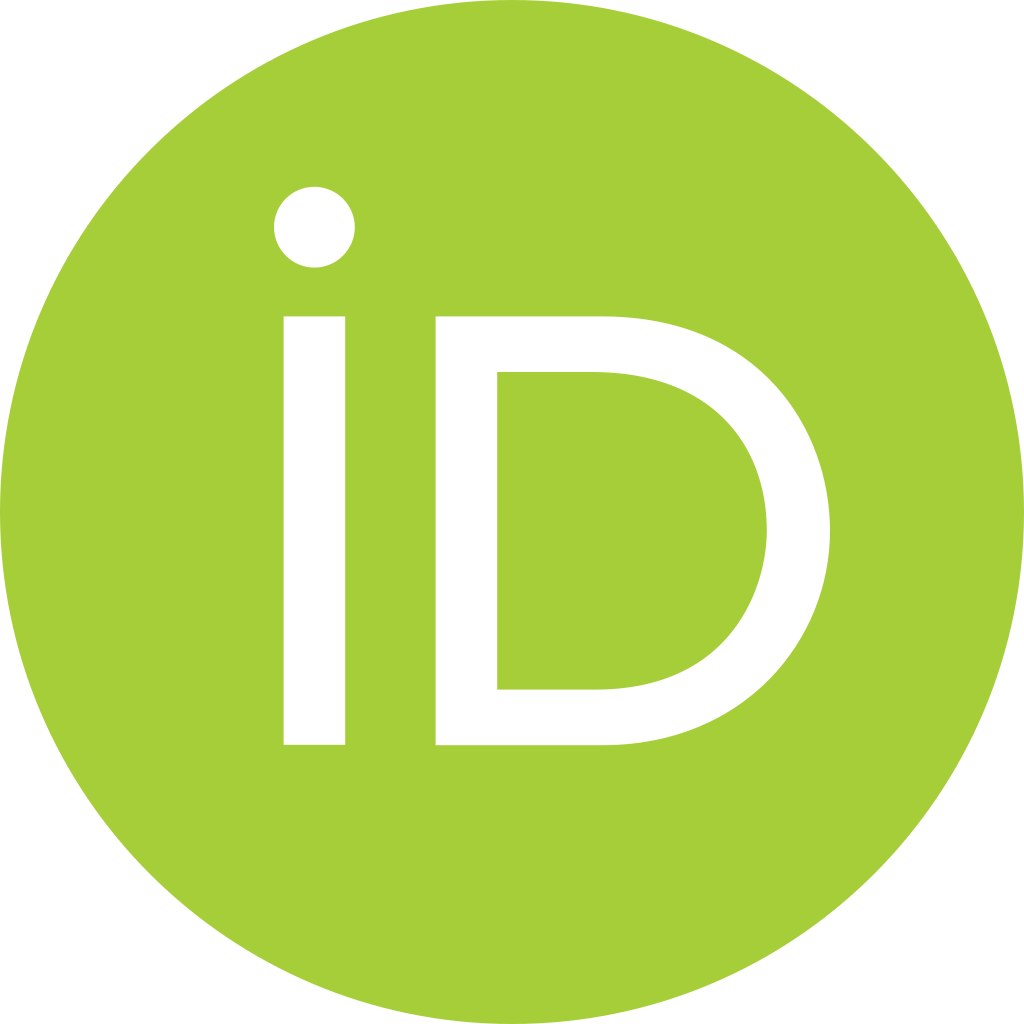}}
\and
Harith Alani\inst{3}\href{https://orcid.org/0000-0003-2784-349X}{\includegraphics[scale=0.009]{orcid.png}} 
\and
Eirini Ntoutsi\inst{1,2}\href{https://orcid.org/0000-0001-5729-1003}{\includegraphics[scale=0.009]{orcid.png}}}
\authorrunning{S. Ghodsi et al.}
\institute{
L3S Research Center, Leibniz Universit\"at Hannover, Germany
\email{ghodsi@l3s.de}\and
Freie Universit\"at Berlin, Dept. of Math. and Comp. Science, Germany
\email{eirini.ntoutsi@fu-berlin.de}\and
The Open University,
Milton Keynes, United Kingdom
\email{h.alani@open.ac.uk}
}

\maketitle              % typeset the header of the contribution
\begin{abstract}
With the ever growing involvement of data-driven AI-based decision making technologies in our daily social lives, the fairness of these systems is becoming a crucial phenomenon. However, an important and often challenging aspect in utilizing such systems is to distinguish validity for the range of their application especially under distribution shifts, i.e., when a model is deployed on data with different distribution than the training set. 
In this paper, we present a case study on the newly released American Census datasets, a reconstruction of the popular Adult dataset, to illustrate the importance of context for fairness and show how remarkably can spatial distribution shifts affect predictive- and fairness-related performance of a model. The problem persists for fairness-aware learning models with 
the effects of context-specific fairness interventions differing across the states and different population groups. 
Our study suggests that robustness to distribution shifts is necessary before deploying a model to another context.

\keywords{Fairness-aware Learning \and Distribution Shifts \and  Responsible AI}

\end{abstract}

\section{Introduction}\label{Intro}
\label{sec:Intro}
\input{Intro}

\section{Related works}
\label{sec:related_works}
\input{related_works}

\section{Preliminaries and research questions}
\label{sec:preliminaries}

\subsection{Dataset}\label{Dataset}
\label{sec:Dataset}
\input{Dataset}

\subsection{Fairness notions}
\label{sec:Fairness_notions}
\input{Fairness_notions}

%-----------
\subsection{Research questions}
\label{sec:RQs}
\input{RQs}

%-----------

\section{Experiments} 
\label{sec:EXP}
\input{EXP}

\subsection{Local vs Global model}
\label{exp:LocalGlobalModels}
\input{exp_part_1}

\subsection{Impact of local interventions for fairness to spatial generalization}
\label{exp:fair_models}
\input{exp_part_2}

\subsection{Context similarity}
\label{exp:ContextSimilarity}
\input{exp_part_3}

\section{Conclusion}
\label{sec:conclusion}
\input{Conclusion}

\section*{Acknowledgment}
\input{NoBIAS}

\bibliographystyle{splncs04}
\bibliography{References}

\end{document}

%% file: Intro.tex
% Introduction
Recent years have brought extraordinary advances in the field of Artificial Intelligence (AI). AI now replaces humans at many critical decision points, such as who will get a loan~\cite{DBLP:conf/icse/VermaR18} and who will get hired for a job~\cite{DBLP:journals/widm/NtoutsiFGINVRTP20}.
There are clear benefits to algorithmic decision-making; unlike people, machines do not become tired or bored {\cite{DBLP:journals/csur/MehrabiMSLG21}}, and can take into account orders of magnitude more factors than people can. However, like people, data-driven algorithms are vulnerable to biases that render their decisions “unfair”. In the automated decision-making, fairness is the absence of any prejudice or favoritism toward an individual or a group based on their inherent or acquired protected attributes such as `race' or `gender'. Thus, an unfair algorithm is one whose decisions are skewed toward a particular group of people.

The importance of context for fairness especially under distribution shift is well recognized by the AI/ML (short for Machine Learning) community~\cite{DBLP:conf/aaai/RezaeiLMZ21,DBLP:conf/aies/BiswasM21,DBLP:conf/fat/SinghSMC21}. Practically, unfairness under distribution shift occurs when training and testing distributions differ. 

Still, it is a common practice in ML to
learn a model over a given dataset and deploy it in the real world ignoring often distribution shifts between the training data and the target application data. This may lead to predictive performance degradation. The distribution shift also affects the discriminatory performance of the model; a fair model during training might turn into a discriminatory one during deployment.

The main goal of this paper is to empirically study such discrepancies due to distribution shifts between model training and model deployment. For our study, we use the recently released US Census dataset~\cite{DBLP:conf/nips/DingHMS21}, which comprises a reconstruction of the popular Adult dataset~\cite{Dua:2019}, and study how the spatial context (i.e., the location/state of the collected data) affects predictive- and fairness-related performance.
In particular, we focus on the binary income prediction task using \emph{``Race''} as the protected attribute and analyze the role of spatial context in the model performance for the different population subgroups for vanilla and fairness-aware models. Our findings show that blindly deployment of a model on a different context/state might result in performance degradation, which is higher for some subgroups leading to unfairness.
We also compare the performance of a global model, trained upon the whole US, with local/context-specific models. The results show that the global model performs better during deployment comparing to the local ones, still the degradation in its performance differs across the subgroups.

The rest of the paper is organized as follows: 
In Section \ref{sec:preliminaries}, we describe basic concepts, the dataset and our research questions. The most recent researches working with the subject dataset are reviewed in Section~\ref{sec:related_works}. 
The experimental findings are provided in Section~\ref{sec:EXP}.  Section~\ref{sec:conclusion} concludes this work and points out to future directions.

%% file: related_works.tex
The newly released US Census dataset \cite{DBLP:conf/nips/DingHMS21}, that we will call \textit{new Adult datasets} hereafter, is intended to replace the \textit{Adult} dataset \cite{Dua:2019}, one of the most popular datasets in the domain of fairness-aware ML~\cite{DBLP:journals/widm/QuyRIZN22}.
Despite its recency, the dataset has attracted a lot of interest from the community due to the variety of tasks and its span.
In \cite{DBLP:journals/corr/abs-2205-04610} three research questions; namely \textit{label constitution, subgroup formation, and moving beyond existing metrics} are asked for bulding a fair-learning pipeline using intersectional fairness. The questions are tackled by providing domain knowledge, studying over data imbalance strategies and introducing methodologies for utilizing intersectionality in fairness-aware learning. The new Adult data is represented as intersection of binary protected attributes. Despite the important step toward intersectionality, it lacks the distribution shift effect. \cite{diciccio2022predictive} introduces statistical ways of estimating predictive rate parity and also proposes a multi-objective post-processing mitigation approach and further tests validity of their proposal using the new Adult datasets on data from Texas state.  

\begin{figure*}[!htb]%
    \centering
    \centerline{\includegraphics[width=1\textwidth, height=\textheight]{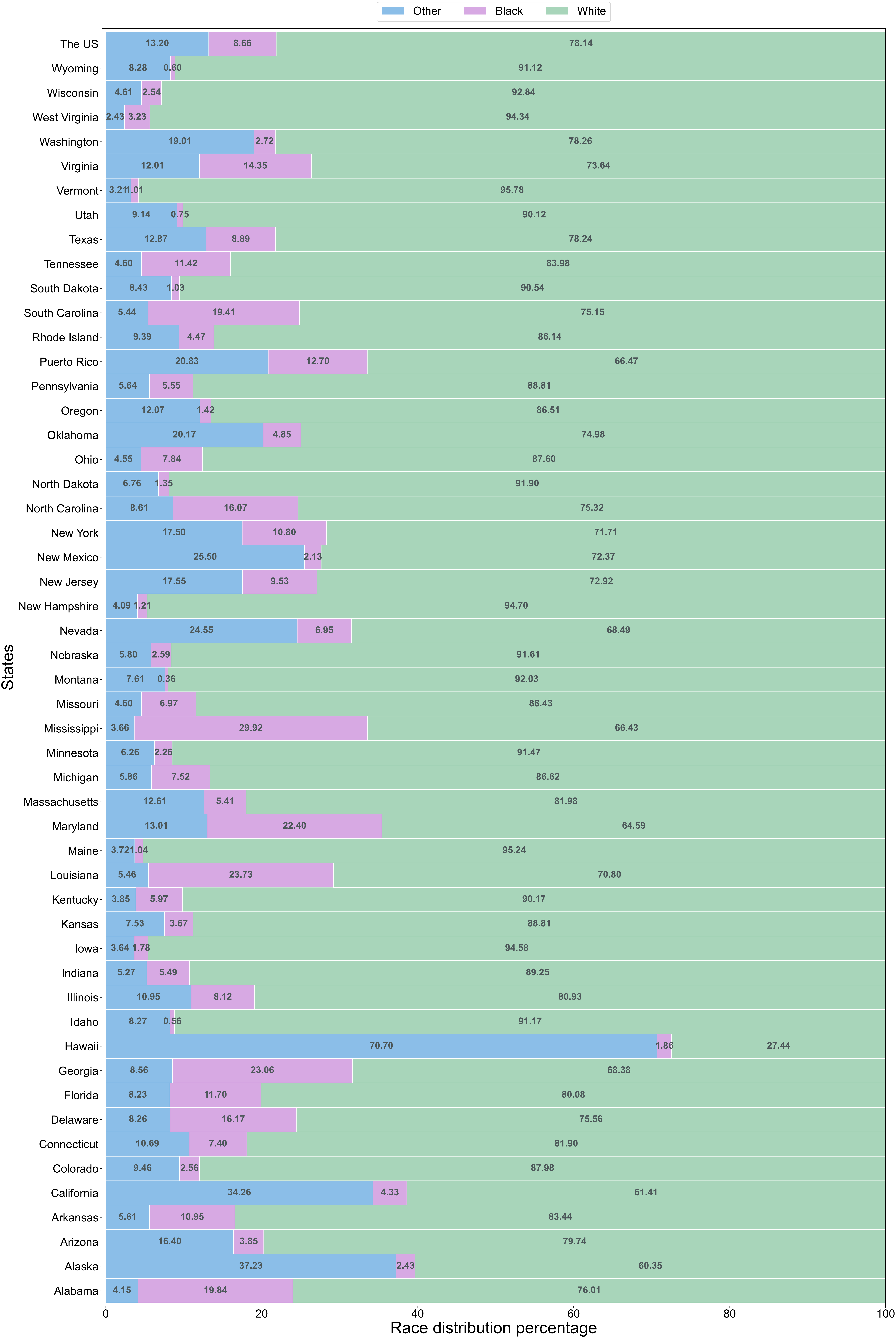}}%
    \caption{Percentage of racial groups (categories: \textit{\{White, Black, Other\}}) per state, incl. US.}
    \label{fig: racialDistribution}
\end{figure*}

%\ inputting the table dataset 
\input{tab2}
\label{Table_dataset}

\cite{DBLP:journals/corr/abs-2107-04423} proposes a method to train proxies that give acceptable approximations of missing protected attributes. They test their method using 5 different tasks of the new datasets but only for one state (New York). Both of the above methods lack the spatial generalizabilty and again are ignoring the effect of out-of-distribution deployment. Moreover, they use datasets with binary protected attributes. 
In \cite{DBLP:journals/corr/abs-2202-01147} a calibrating subset selection strategy is proposed, claiming to develop a distribution-free monitoring scheme. However, their analysis also suffers from dependence on binary protected attribute.

The data is also studied in other domains of ML research. For example in \cite{DBLP:journals/corr/abs-2206-01067} it is used for validating a conformal prediction method for sequential prediction. Also it has been reviewed in a number of surveys including \cite{DBLP:journals/corr/abs-2202-01711,DBLP:journals/corr/abs-2102-05242,DBLP:journals/widm/NtoutsiFGINVRTP20} so far.

%% file: tab2.tex
 \begingroup
 \begin{table*}%[!htb]%
    \tiny
    \caption {\label{table:dataset_details}Statistics for the different states incl. US (for 2019)}
\begin{adjustbox}{width=\textwidth,totalheight=\textheight}
%\scalebox{0.8}{
%\resizebox*{!}{\textheight}{ 
\begin{tabularx}{\linewidth} {lcclllllllc}
%{@{}l*{10}{l}l@{}}
    \toprule\toprule
    &&&&&&&
    \multicolumn{3}{c}{\bf{Race Distribution}} \\
    \cmidrule(r){8-10}
    \bf{States} & 
    \bf{Abr} & 
    \bf{\thead{Target\\ (Task)}} & 
    \bf{\thead{\#Attribs\\(all,task)}} &
    \bf{\thead{Protected\\Attributes}} &
    \bf{\thead{\#Sample}} &
    \bf{\thead{\#Sample\\(cleaned)}}
    & \bf{\thead{White\\ (rate)}}
    & \bf{\thead{Black\\ (rate)}}
    & \bf{\thead{Other\\ (rate)}}
    &
    \bf{\thead{Class\\ratio (+:-)}} 
    \\
    \midrule

    Alabama        & AL & bin(Inc) & 288 , 10 & R,S,A,MS & 48,928 & 22,798   & \bf{76.01\%}  & 19.84\%  & 4.15\%   & 1:2.01 \\
    
    Alaska         & AK & bin(Inc) & 288 , 10 & R,S,A,MS & 6,663  &  3,417   & \bf{60.35\%}  & 2.43\%   & 37.23\%   & 1:1.4 \\
    
    Arizona        & AZ & bin(Inc) & 288 , 10 & R,S,A,MS & 70,872 & 34178    & \bf{79.74\%}  & 3.85\%   & 16.40\%   & 1:1.75 \\
    
    Arkansas       & AR & bin(Inc) & 288 , 10 & R,S,A,MS & 30,403 & 14,104   & \bf{83.44\%}  & 10.95\%  & 5.61\%   & 1:2.55  \\
    
    California     & CA & bin(Inc) & 288 , 10 & R,S,A,MS & 380,091 & 197,193 & \bf{61.41\%}  & 4.33\%   & 34.26\%   & 1:1.28  \\
    
    Colorado       & CO & bin(Inc) & 288 , 10 & R,S,A,MS & 57,142 & 32,264   & \bf{87.98\%}  & 2.56\%   & 9.46\%    & 1:1.26  \\ 
    
    Connecticut    & CT & bin(Inc) & 288 , 10 & R,S,A,MS & 35,578 & 19,872   & \bf{81.90\%}  & 7.4\%    & 10.69\%   & 1:1.02  \\
    
    Delaware       & DE & bin(Inc) & 288 , 10 & R,S,A,MS &  9,261  &  4,755  & \bf{75.56\%}  & 16.17\%  & 8.26\%    & 1:1.32 \\ 
    
    Florida        & FL & bin(Inc) & 288 , 10 & R,S,A,MS & 205,294 & 100,426 & \bf{80.08\%}  & 11.7\%   & 8.23\%    & 1:1.84  \\ 
    
    Georgia        & GA & bin(Inc) & 288 , 10 & R,S,A,MS & 101,652 & 51,032  & \bf{68.38\%}  & 23.06\%  & 8.56\%    & 1:1.67 \\
    
    Hawaii         & HI & bin(Inc) & 288 , 10 & R,S,A,MS & 14,228  &  7624   & 27.44\%   & 1.86\%   & \bf{70.7\%}   & 1:1.39 \\ 
    
    Idaho          & ID & bin(Inc) & 288 , 10 & R,S,A,MS &  17,360 &  8,564  & \bf{91.17\%}  & 0.56\%   & 8.27\%    & 1:2.3  \\
    
    Illinois       & IL & bin(Inc) & 288 , 10 & R,S,A,MS & 125,007 & 66,051  & \bf{80.93\%}  & 8.12\%   & 10.95\%   & 1:1.38  \\ 
    
    Indiana        & IN & bin(Inc) & 288 , 10 & R,S,A,MS & 67,445 &  35,222  & \bf{89.25\%}  & 5.49\%   & 5.27\%    & 1:2.05  \\ 
    
    Iowa           & IA & bin(Inc) & 288 , 10 & R,S,A,MS & 32,721 &  17,620  & \bf{94.58\%}  & 1.78\%   & 3.64\%    & 1:1.96  \\ 

    Kansas         & KS & bin(Inc) & 288 , 10 & R,S,A,MS & 29,697 &  15,793  & \bf{88.81\%}  & 3.67\%   & 7.53\%    & 1:1.95  \\ 
    
    Kentucky       & KY & bin(Inc) & 288 , 10 & R,S,A,MS & 45,649 &  22,003  & \bf{90.17\%}  & 5.97\%   & 3.85\%    & 1:1.99  \\
    
    Louisiana      & LA & bin(Inc) & 288 , 10 & R,S,A,MS & 43,896 &  20,897  & \bf{70.80\%}  & 23.73\%  & 5.46\%    & 1:1.78   \\ 
    
    Maine          & ME & bin(Inc) & 288 , 10 & R,S,A,MS & 13,493 &  7,104   & \bf{95.24\%}  & 1.04\%   & 3.72\%    & 1:2.15 \\
    
    Maryland       & MD & bin(Inc) & 288 , 10 & R,S,A,MS & 60,237 &  32,865  & \bf{64.59\%}  & 22.4\%   & 13.01\%   & 1.05:1  \\
    
    Massachusetts  & MA & bin(Inc) & 288 , 10 & R,S,A,MS & 70,579 &  40,639  & \bf{81.98\%}  & 5.41\%   & 12.61\%   & 1:1.01   \\
    
    Michigan       & MI & bin(Inc) & 288 , 10 & R,S,A,MS & 100,078 & 50,475  & \bf{86.62\%}  & 7.52\%   & 5.86\%    & 1:1.76  \\
    
    Minnesota      & MN & bin(Inc) & 288 , 10 & R,S,A,MS & 56,670 &  31,243  & \bf{91.47\%}  & 2.26\%   & 6.26\%    & 1:1.45  \\

    Mississippi    & MS & bin(Inc) & 288 , 10 & R,S,A,MS & 29,217 &  13,159  & \bf{66.43\%}  & 29.92\%  & 3.66\%   & 1:2.51  \\
    
    Missouri       & MO & bin(Inc) & 288 , 10 & R,S,A,MS & 63,174 &  31,589  & \bf{88.43\%}  & 6.97\%   & 4.6\%   & 1:1.96  \\
    
    Montana        & MT & bin(Inc) & 288 , 10 & R,S,A,MS & 10,649 &  5,547   & \bf{92.03\%}  & 0.36\%   & 7.61\%   & 1:2.09  \\ 
    
    Nebraska       & NE & bin(Inc) & 288 , 10 & R,S,A,MS & 19,766 &  10,816  & \bf{91.61\%}  & 2.59\%   & 5.8\%   & 1:1.94  \\
    
    Nevada         & NV & bin(Inc) & 288 , 10 & R,S,A,MS & 29,347 &  14,915  & \bf{68.49\%}  & 6.95\%   & 24.55\%   & 1:1.84 \\
   
    New Hampshire  & NH & bin(Inc) & 288 , 10 & R,S,A,MS & 13,896 &  7,903   & \bf{94.70\%}  & 1.21\%   & 4.09\%   & 1:1.29  \\
    
    New Jersey     & NJ & bin(Inc) & 288 , 10 & R,S,A,MS & 88,459 &  48,355  & \bf{72.92\%}  & 9.53\%   & 17.55\%   & 1.05:1  \\
    
    New Mexico     & NM & bin(Inc) & 288 , 10 & R,S,A,MS & 19,281 &  9,027   & \bf{72.37\%}  & 2.13\%   & 25.5\%   & 1:2  \\
    
    New York       & NY & bin(Inc) & 288 , 10 & R,S,A,MS & 198,788 & 104,330 & \bf{71.71\%}  & 10.8\%   & 17.5\%   & 1:1.25  \\
    
    North Carolina & NC & bin(Inc) & 288 , 10 & R,S,A,MS & 103,516 & 52,041  & \bf{75.32\%}  & 16.07\%  & 8.61\%   & 1:1.86  \\
    
    North Dakota   & ND & bin(Inc) & 288 , 10 & R,S,A,MS & 7,960  &  4,455   & \bf{91.90\%}  & 1.35\%   & 6.76\%   & 1:2.2  \\ 
    
    Ohio           & OH & bin(Inc) & 288 , 10 & R,S,A,MS & 119,589 & 62,102  & \bf{87.60\%}  & 7.84\%   & 4.55\%   & 1:1.78  \\ 
    
    Oklahoma       & OK & bin(Inc) & 288 , 10 & R,S,A,MS & 37,792 &  17,790  & \bf{74.98\%}  & 4.85\%   & 20.17\%  & 1:1.64 \\
    
    Oregon         & OR & bin(Inc) & 288 , 10 & R,S,A,MS & 42,080 &  21,876  & \bf{86.51\%}  & 1.42\%   & 12.07\%  & 1:1.58 \\
    
    Pennsylvania   & PA & bin(Inc) & 288 , 10 & R,S,A,MS & 129,194 & 68,002  & \bf{88.81\%}  & 5.55\%   & 5.64\%   & 1:1.6  \\

    Puerto Rico    & PR & bin(Inc) & 288 , 10 & R,S,A,MS & 27,820 &  9,270   & \bf{66.47\%}  & 12.7\%   & 20.83\%  & 1:7.12  \\
    
    Rhode Island   & RI & bin(Inc) & 288 , 10 & R,S,A,MS & 10,447 &  5,930   & \bf{86.14\%}  & 4.47\%   & 9.39\%   & 1:1.29 \\
    
    South Carolina & SC & bin(Inc) & 288 , 10 & R,S,A,MS & 50,893 &  25,085  & \bf{75.15\%}  & 19.41\%  & 5.44\%   & 1:1.96  \\
    
    South Dakota   & SD & bin(Inc) & 288 , 10 & R,S,A,MS & 9,128  &  4,949   & \bf{90.54\%}  & 1.03\%   & 8.43\%   & 1:2.17  \\
    
    Tennessee      & TN & bin(Inc) & 288 , 10 & R,S,A,MS & 68,415 &  34,286  & \bf{83.98\%}  & 11.42\%  & 4.6\%    & 1:1.96  \\ 
    
    Texas          & TX & bin(Inc) & 288 , 10 & R,S,A,MS & 272,776 & 138,293 & \bf{78.24\%}  & 8.89\%   & 12.87\%  & 1:1.55   \\
    
    Utah           & UT & bin(Inc) & 288 , 10 & R,S,A,MS & 32,371 &  16,774  & \bf{90.12\%}  & 0.75\%   & 9.14\%   & 1:1.81  \\
    
    Vermont        & VT & bin(Inc) & 288 , 10 & R,S,A,MS & 6,543  &  3,770   & \bf{95.78\%}  & 1.01\%   & 3.21\%   & 1:1.78    \\
    
    Virginia       & VA & bin(Inc) & 288 , 10 & R,S,A,MS & 84,938 &  46,107  & \bf{73.64\%}  & 14.35\%  & 12.01\%  & 1:1.19    \\

    Washington     & WA & bin(Inc) & 288 , 10 & R,S,A,MS & 77,879 &  40,956  & \bf{78.26\%}  & 2.72\%   & 19.01\%  & 1:1.21   \\
    
    West Virginia  & WV & bin(Inc) & 288 , 10 & R,S,A,MS & 18,130 &  8,152   & \bf{94.34\%}  & 3.23\%   & 2.43\%   & 1:2.18   \\
    
    Wisconsin      & WI & bin(Inc) & 288 , 10 & R,S,A,MS & 59,767 &  32,466  & \bf{92.84\%}  & 2.54\%   & 4.61\%   & 1:1.74   \\
    
    Wyoming        & WY & bin(Inc) & 288 , 10 & R,S,A,MS & 5,967  &  3,154   & \bf{91.12\%}  & 0.6\%    & 8.28\%   & 1:1.69   \\
    
    \midrule
    
    The US  & US & bin(Inc) & 288 , 10 & R,S,A,MS & 3,239,553 & 1,672,300   & \makecell{1,306,709\\(\bf{78.14\%})}  & \makecell{144,781\\(8.66\%)}      & \makecell{220,810\\(13.2\%)}   & 1:1.52 \\
    
    \bottomrule\bottomrule
\end{tabularx}
%}
%}
\end{adjustbox}
\end{table*}
\endgroup
\clearpage

%% file: Dataset.tex
The dataset ~\cite{DBLP:conf/nips/DingHMS21} provides census information including demographics, economical,
and working status of US citizens. With a span of over 20 years it allows for research in temporal and spatial distribution shifts and their effect on fairness. The feature space consists of 286 features (from which only 10 are relevant in our analysis) and a class feature. It includes 5 tasks for predicting \emph{income}, \emph{public health insurance}, \emph{residential address}, \emph{employment} and \emph{commuting time to workplace} can be from which we focus on the \emph{income prediction task} and study only the spatial context to examine the connection between geographical shifts and bias. Therefore, we focus only on the latest year of available data, \emph{2019}.
The datasets provide actual income of individuals. Similar to~\cite{DBLP:conf/nips/DingHMS21}, we turn this into a binary classification class to predict whether an individual earns an income of more than 50k:  $income \in  \{\le50K, >50K\}$, the positive class being $``>50K''$.

We pre-process the data following the guidelines of~\cite{DBLP:conf/nips/DingHMS21}. The filtering applied to some of the features (including the class feature) are: age (greater than 16), income (greater than 100), and working hours per week (greater than 0).  
In total, there are \#51 datasets (1 for each state, we call them \emph{local datasets} or \emph{context-specific datasets}). Moreover, a dataset for the whole US is provided, excluding the ``Puerto Rico'' island; we refer to it as the \emph{global-dataset} or \emph{US-dataset}.
A summary of the total \#52 datasets is given in Table~ \ref{table:dataset_details}.
The most populated state is California, followed by New York, Texas and Florida. The least populated state is Wayoming.
The most class balanced state is Massachusetts (1:1:01) the least balanced is Arkansas (1:2.55). In the overall US (dataset ``The US'') the imbalance ratio is 1:1.52.

\subsubsection{Spatial race distribution}
We consider ``race'' as the protected attribute; original feature name RAC1P (Recoded detailed race code). The value domain consists of 9 values:  \{``White alone'', ``Black or African American alone'', ``American Indian alone'', ``Alaska Native alone'', ``American Indian and Alaska Native tribes specified, or American Indian or Alaska Native, not specified and no other races'', ``Asian alone'', ``Native Hawaiian and Other Pacific Islander alone'', ``Some Other Race alone'', ``Two or More Races''\}. 
Due to the small representation of the last seven categories, we map them in a bigger group called {\emph{``Other''}. So, the race attribute has 3 groups: \textit{White, Black, and Other}}. We believe studying bias w.r.t. race for these datasets can be more challenging than taking for instance gender as protected attribute. The reason is because of highly imbalanced distribution of racial groups over the states while gender and age are much smoother.

To the best of our knowledge, this is the first work studying ``Race'' from the new Adult datasets as a non-binary protected attribute.
Fig. \ref{fig: racialDistribution} illustrates the racial distribution across the different states including US. \emph{White} is the dominant group in all states except for Hawaii that is dominated by the \emph{``Other''} group (70,70\%).
Vermont has the highest ratio of \emph{``White''} inhabitants (95.78\%) and subsequently one of the lowest black populations. However, the lowest \emph{``Black''} community belongs to Montana with 0.36\% and only 20 samples after cleaning. Missisipi has the largest ratio of Black population (66,43\%).

%% file: Fairness_notions.tex
In this paper, we adopt the \textit{Equalized Odds} (Eq.Odds for short) \cite{DBLP:conf/cikm/IosifidisN19} definition of fairness which measures the difference in prediction errors between the protected and non-protected groups for both classes as: 
$\left|{\delta}FPR \right| + \left| {\delta}FNR\right|$.

$\delta$FNR measure a.k.a the \emph{Equal Opportunity}, is the difference of the probability of subjects from both the protected and unprotected groups that belong to the \textbf{positive} class to have a negative predictive value~\cite{DBLP:conf/icse/VermaR18}: $${\delta}FNR=\left| {P(\hat{Y} = 0|Y = 1, g = w) - P(\hat{Y}=0|Y = 1, g = b/o)}\right| $$ where $\hat{Y}$ is the predicted label, Y is the actual label and $g \in G = \{w, b, o\}$ is the protected attribute. 
In this study, ``w'' stands for the non-protected group and the other two show protected.
A classifier satisfies the equality of opportunity if its group-wise $\delta$FNR is zero. 
Similarly, $\delta$FPR a.k.a. as \textit{Predictive Equality} is the difference of the probability of subjects from both the protected and unprotected that belong to the \textbf{negative} class to have a positive predictive value:
 $$\delta FPR=\left| {P(\hat{Y}= 1|Y = 0, g = w) - P(\hat{Y} = 1|Y = 0, g = b/o)}\right| $$

The value range for each of $\delta$FNR and $\delta$FPR is [0,1], where 0 stands for no discrimination and 1 stands for maximum discrimination.

%% file: RQs.tex
For our study, we form the following research questions:

\noindent \textbf{RQ1. 
Local vs Global model:} \emph{How local models learned from particular states compare to a global model trained upon data from the whole US, w.r.t both predictive and fairness-related performance?}

\noindent \textbf{RQ2. 
Impact of local interventions for fairness to spatial generalization:} 
\emph{How correcting for the unfairness of a local model, affects spatial generalization, w.r.t both predictive and fairness-related performance?}

\noindent \textbf{RQ3. 
Using context similarity to understand spatial differences:} 
\emph{How can we detect context similarity, i.e., similar states, which can be used to predict how a model will perform to a different context/state, w.r.t both predictive and fairness-related performance?}

%% file: EXP.tex
The goal of our experimental study is to study the effect of (spatial) context in model performance and address the research questions asked previously in section \ref{sec:RQs}. To this end, we first (Section~\ref{exp:LocalGlobalModels}) analyze the effect of spatial distribution shifts by training vanilla LR models locally and globally in a particular state/context and deploying them to other states/contexts (\textbf{RQ1}). The second experiment (Section~\ref{exp:fair_models}) will follow a similar process as the first one w.r.t. a fair-LR classifier but this time with pre-processing and in-processing embodied fairness interventions to address \textbf{RQ2}. Finally, by studying context similarity of states data we will analyze the similarity/dissimilarity of performances by statistically comparing corresponding data distributions in Section~\ref{exp:ContextSimilarity} which answers \textbf{RQ3}.

In terms of fairness performance, we report mainly on the  difference of false classification rates ($\delta$FPR and $\delta$FNR) between the privileged groups (W) and the other two racial groups (B, O). We also look at Eq.Odds measure which aggregates the $\delta$FPR and $\delta$FNR.

Throughout the experiments, we consider \emph{race} as the protected attribute. In this study, we consider white (``W'') group as the privileged and the ``B, O'' as the unprivileged groups. The reason is because the ``W'' group is the dominant group in all the states (Figure~\ref{fig: racialDistribution}) except for Hawaii.

The base model is a Linear Regression model (LR). The experimental analysis are based on stratified 10-fold cross validation technique to ensure reliability of the results. The experiments include an in-distribution (splitting the dataset to train/test sets) and an out-of-distribution (train on a state-data and testing on the other 50 context/states) deployment. In the out-of-distribution global testing the deployment set is excluded from the training set to ensure that training and testing sets are disjoint.

The datasets and problem formulation are the same as what the authors of \cite{DBLP:conf/nips/DingHMS21} use. The two (pre/in-processing) fairness methods (used later in Section~\ref{exp:fair_models}) are adopted from the \href{https://github.com/Trusted-AI/AIF360}{IBM AIF360} framework \cite{AIF360}.

%% file: exp_part_1.tex
% exp 1

\begin{figure*} [!htb]
    \begin{subfigure}{0.48\linewidth}
        \centering
        \includegraphics[width=\linewidth]{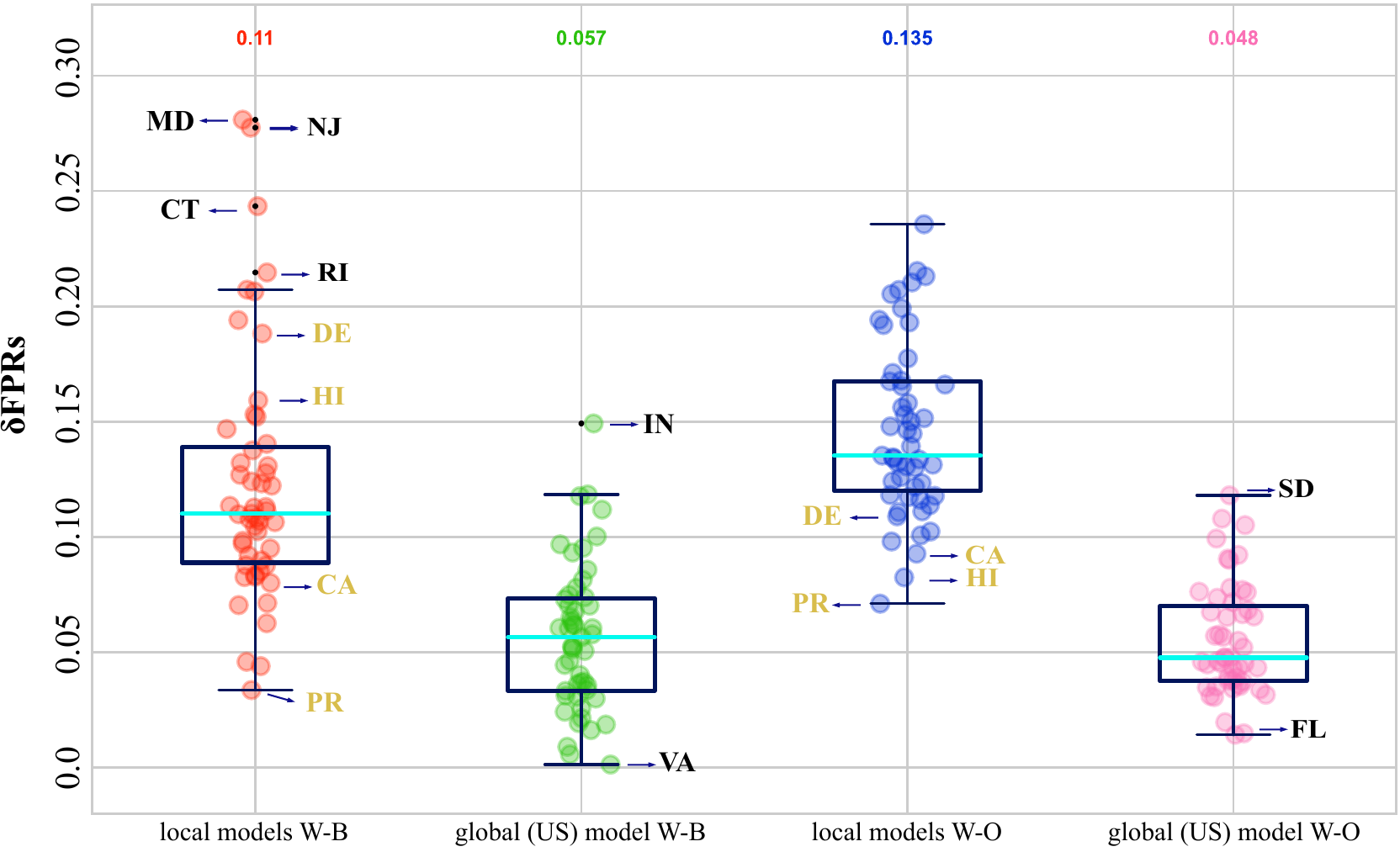} 
    \end{subfigure}
    \begin{subfigure}{0.48\linewidth}
        \centering
        \includegraphics[width=\linewidth]{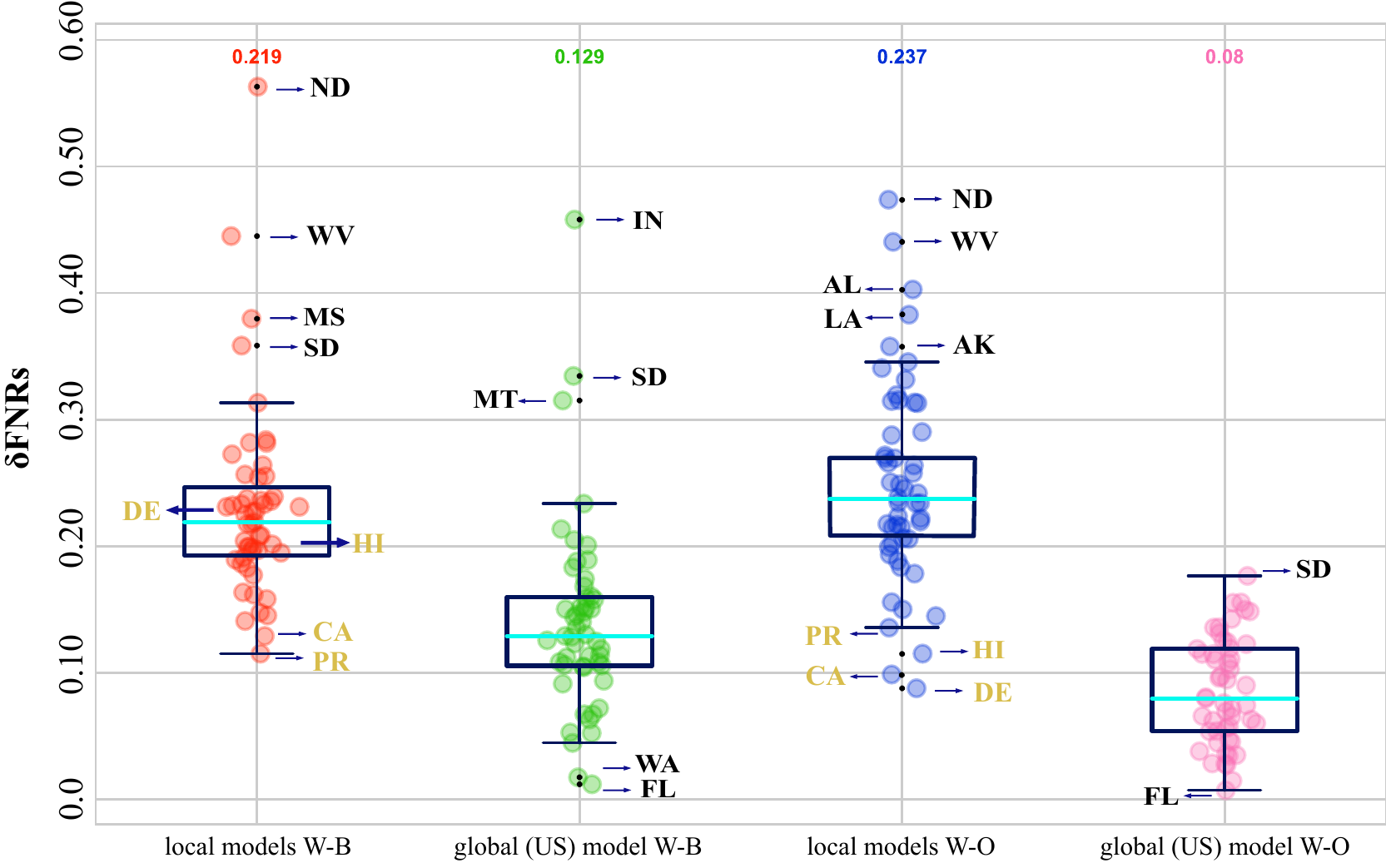} 
    \end{subfigure}
    
    \caption{Spatial distribution shifts effects on fairness - Vanilla models case: local and global models are evaluated in terms of $\delta FPR$ (a) and $\delta FNR$ (b) scores between W-B and W-O subroups.
    Each point on the local models boxplot correspond to the median of a state-model deployed to the other states.}
    \label{Fig: bp_n2}
    \vspace{-5mm}
\end{figure*}

\paragraph{\textbf{Vanilla models}}
For the local models, we train a (vanilla) LR on each state and deploy them to the other 50 states.
For the global model, we train a vanilla LR on the US-data and deploy it to all states.

We report on the fairness-related behavior of the local and global  models. In particular, we report on the $\delta FPR$ and $\delta FNR$ scores between ``White and Black'' (short W-B) subgroups and ``White and Other'' (short W-O) subgroups. 
The results are shown in Figure~\ref{Fig: bp_n2}. Each point in the local models boxplots, is the median of out-of-distribution deployment results of that specific model. The light blue lines represent the median of boxplots, which in case of local models are the medians of the medians. The median values of each chart is also shown on top of the charts.

Overall, the local models perform worse than the global model, for both W-B and W-O subgroup comparisons. Specifically, the medians and the range of boxplots (indicating the variance) report higher $\delta$FNR and $\delta$FPR measures compared to the ones of the global-model.
Results show that there are states with medians comparable to the median of the global model. Specifically, the ones between the min and the first quartile of the local boxplots like Delaware, California, and Hawaii annotated on Figure~\ref{Fig: bp_n2}.

Comparing the performance differences across different subgroups, for the global model the difference in discrimination scores is higher regarding the W-B subgroups compared to the W-O. Conversely, the W-O difference is the dominant discriminatory behavior of the local models for both ($\delta$FPR and $\delta$FNR) measures. The variance of W-B is slightly higher than the W-O subgroups in both the local and the global models. 

The highest median comes from the $\delta$FNR of the local models W-O, probably also due to the many outlier points. 
The top-3 outliers correspond to models of \emph{Delaware, California, and Hawaii} with medians of 0.089, 0.096, and 0.011 (for the W-O $\delta$FNR) respectively. These states along with Puerto Rico (the best within quartiles), have also a similar trend in the $\delta$FPR chart. They fall within the min and the first quartile of the blue boxplot in the left. They are highlighted with Khaki color on the charts. 

The state Puerto Rico has a suspicious performance. Its good performance in $\delta$FNR is due to ignoring the positive instances in the protected groups and miss-classifying them to the negative class. Note that, this state has the lowest ratio of positive instances ($\frac{1}{(1+7.12)}=12\%$) according to Table~\ref{table:dataset_details}.  
Excluding PR, there seem to be a correlation to the high ratio of ``Other'' subgroup for the remaining three states and their smaller discrimination toward the Other group. However, only the California model repeats the same behavior (good results) for the W-B difference having 0.081, 0.128 regarding the $\delta$FPR and the $\delta$FNR respectively. The two other states report much higher W-B difference. In addition, according to Table~\ref{fig: racialDistribution}, California has a high number of ``Black'' people among the states that can help the classifier better learn this subgroup.

In general, there exist many outlier points in the local models, indicating the varying effects of distribution shifts.

\begin{figure*} [!htb]
    \begin{subfigure}{0.48\linewidth}
        \centering
        \includegraphics[width=\linewidth]{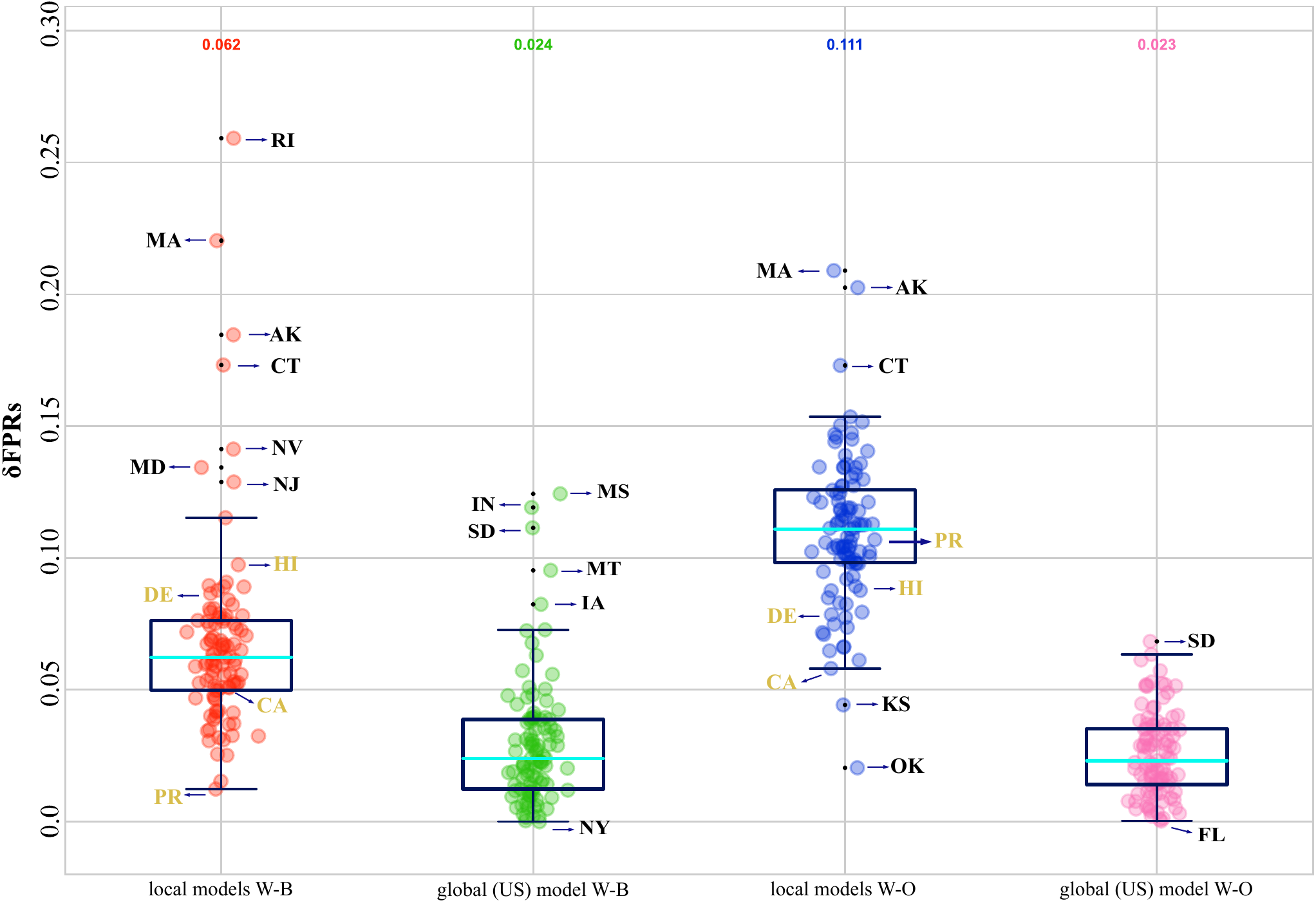} 
        \caption{boxplot of $\delta$FPRs}
    \end{subfigure}
    \begin{subfigure}{0.48\linewidth}
        \centering
        \includegraphics[width=\linewidth]{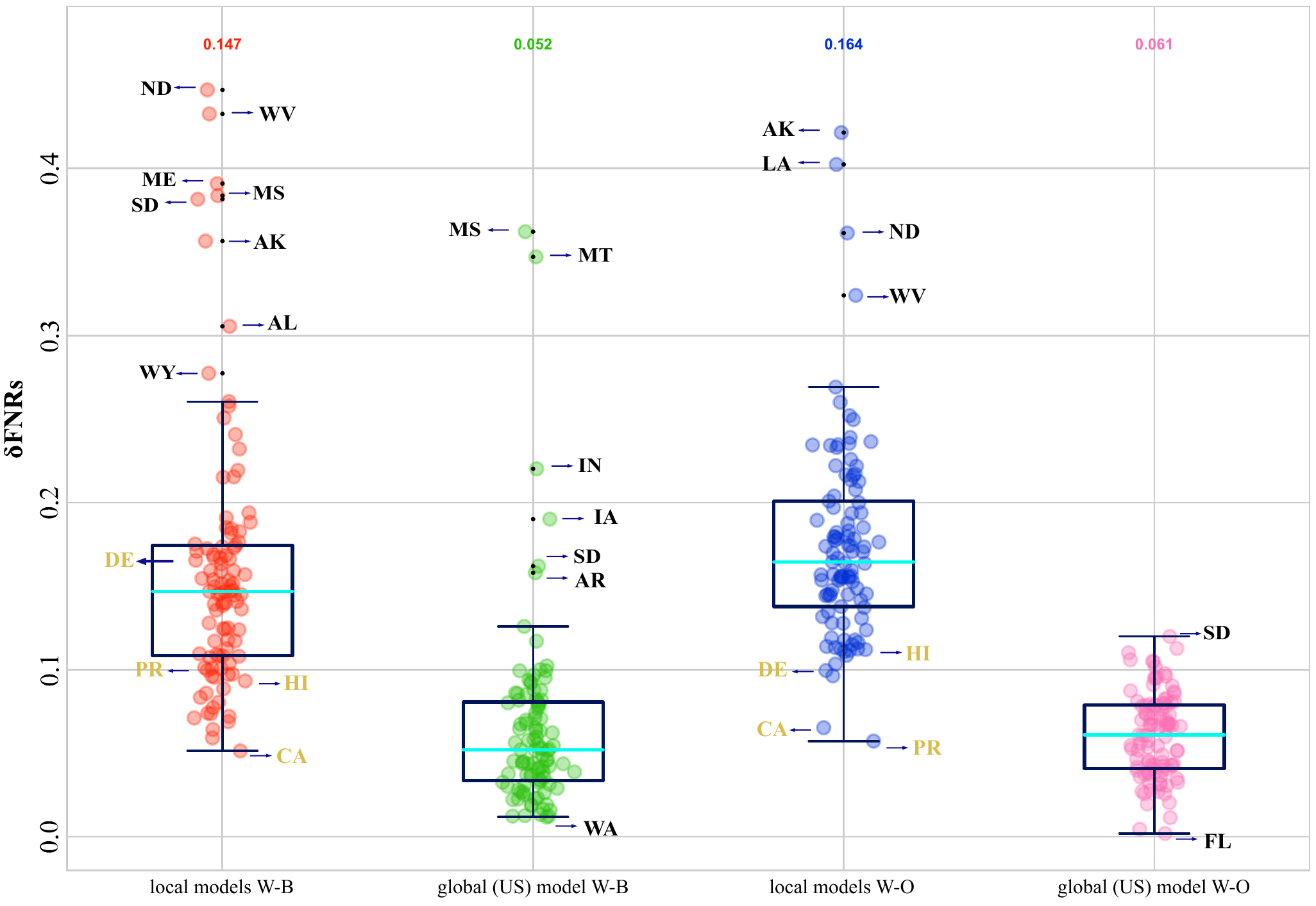} 
        \caption{boxplot of $\delta$FNRs}
    \end{subfigure}
    
    \caption{
    Spatial distribution shifts effects on fairness - Fairness-aware models case: local and global models are evaluated in terms of $\delta FPR$ (a) and $\delta FNR$ (b) scores between W-B and W-O subroups.
    Each point on the local models boxplot correspond to the median of a state-model deployed to the other states.}
    \label{Fig: fair_bp_n2}
    \vspace{-8mm}
\end{figure*}

%% file: exp_part_2.tex
\paragraph{\textbf{Fairness-aware models}}

Next, we evaluate the impact of fairness interventions on the context-specific, i.e., on the state models, as well as on the global model, i.e., the US model under distribution shift. The question is whether a fair-aware model for a particular context/state could retain its fair behavior when applied to other contexts/states. Similarly, for the global model, how a global ``debiased'' model performs across different states.
The experimentation process is similar to the vanilla models case, the only difference is that now the employed model is a fairness-aware model. In particular, our model is a combination of a pre-processing (data level) and in-processing (algorithm level) approach. 
For the pre-processing interventions, we follow the sample re-weighting approach~\cite{DBLP:journals/kais/KamiranC11}, that weights the examples in each protected group and class label combination differently to ensure a fair data representation.
For in-processing interventions we follow the so-called ``prejudice remover'' approach~\cite{DBLP:conf/pkdd/KamishimaAAS12} that adds a regularization term to the LR objective function to enforce classifier’s independence from sensitive information. 

The results are shown in Figure~\ref{Fig: fair_bp_n2}. As expected, the discrimination scores are lower comparing to the vanilla models (Figure~\ref{Fig: bp_n2}) for both global and local models.
Similarly to the vanilla case, the global model results in lower discrimination comparing to the local ones (as seen by the lower median) and moreover the variance of the local models is higher (higher min-max range comparing to the global model). 
There are a few states with smaller $\delta$FPR scores for the W-B subgroups than the global model's median but, they report higher medians than the global model's in the other measures/subgroups.

Comparing the performance across the different subgroups, we see that the local models result into discrimination scores that are higher for the other group (W-O) in comparison to the (W-B) group. The differences are higher comparing to the vanilla case. 
% To be specific, (0.237-0.219 = 0.018) $\approx$ (0.164-0.147=0.017) comparing right charts of Fig2 and Fig3, (0.135-0.11=0.025) < (0.111-0.062=0.049) comparing left chart of Figs 2,3.
The global model behaves more similarly across the subgroups (similar medians), however a closer look at the distribution shows that there are many outliers for the W-B subgroups indicating high discrimination scores for certain states.
They belong to the following states respectively: Missouri, Montana, Indiana, Iowa, South-Dakota, and Arkansas.

\textit{Summary:}
The experiments show that the (spatial) context matters (\textbf{RQ1}); training a (vanilla or fairness-aware) model on a given state and deploying it blindly to other states/contexts is not a good strategy. 
State/Context-specific fairness interventions (\textbf{RQ2}) do not solve the problem and they even amplify the $\delta FPR$ differences between W-O and W-B subgroups.
The global model trained upon representative data from the US, has the fairest (relative) performance compared to the state-wise models. Still, this model differentiates between the different subgroups (it seems to be more fair w.r.t. to the W-O subgroups in the vanilla version; in the fairness-aware version the overall differences are reduced but certain states still score high in discrimination as shown by the many outliers).

%% file: exp_part_3.tex
% exp 3
As already shown, a blind deployment of local models trained on a particular context/state to some other context/state results in performance degradation for some states. Intuitively, this is due to differences in the data distributions. In this section, therefore, we look at similarities and differences between training and deployment states/contexts. To measure context similarity, we use \textit{Maximum mean discrepancy \bf{(MMD)}} \cite{DBLP:journals/jmlr/GrettonBRSS12}, which represents distances between distributions as distances between mean feature embeddings.

Let us suppose we have two datasets \(X\), \(V\) with probability distributions \(P\) and \(Q\). MMD represents the distance between the distributions \(P\) and \(Q\) in terms of L2-distance of feature means (or kernel means) in what's called a Reproducing Kernel Hilbert Space (RKHS): 

\begin{equation}\label{eqn:mmd}
    MMD^{2}(P,Q) = \Vert \mu_{P} - \mu_{Q} \Vert^{2} _{\mathcal{H}}
  \end{equation}
  
Assuming $ \phi: X \rightarrow H $ to be a feature map embedding \(X\) to the embedding space \(H\), we can estimate kernel mean of \(X\) using $ \mu_{P} ~ (\phi(X)) = \frac{1}{n} \sum_{i=1} \phi(x_{i})$. Then Eq~\ref{eqn:mmd} can be rewritten as: 

\begin{equation}\label{eqn:mmd_Fi}
MMD^{2}(P,Q) = \bigg\Vert{\frac{1}{n} \sum_{i=1} \phi(x_{i}) - \frac{1}{m} \sum_{i=1} \phi(v_{i})}\bigg\Vert ^{2} _{\mathcal{H}} 
%- {2 \frac{1}{m.m} \sum_{i} \sum_{j} k(\mathbf{x_{i}}, \mathbf{d_{j}})} + {\frac{1}{m (m-1)} \sum_{i} \sum_{j \neq i} k(\mathbf{d_{i}}, \mathbf{d_{j}})}
\end{equation}

The inner product of feature means of \(X \sim P\) and \(V \sim Q\) can be written in terms of the kernel function such that:

\begin{equation}\label{eqn:phi_kernel}
\langle \mu_{P}~(\phi(X), \mu_{Q}~(\phi(V) \rangle_{\mathcal{H}} = E_{P,Q}~ [\langle \phi(X), \phi(V) \rangle_{\mathcal{H}}] = E_{P,Q}~ [k(X,V)] 
\end{equation}

Then using Eq~\ref{eqn:phi_kernel} we can expand Eq~\ref{eqn:mmd} and rewrite it such that:

\begin{equation}\label{eqn:mmd_kernelize}
MMD^{2}(P,Q) = E_{P}~ [k(X,X)] - 2 E_{P,Q}~ [k(X,V)] + E_{Q}~ [k(V,V)] 
\end{equation}

Finally, expanding the Eq~\ref{eqn:mmd_kernelize} the two sample MMD-test can be calculated by:

\begin{equation}\label{eqn:mmd_final}
\begin{split}
    MMD^{2}(X,V) = {\frac{1}{m (m-1)} \sum_{i} \sum_{j \neq i} k(\mathbf{x_{i}}, \mathbf{x_{j}})} - {2 \frac{1}{m.m} \sum_{i} \sum_{j} k(\mathbf{x_{i}}, \mathbf{v_{j}})} \\ +{\frac{1}{m (m-1)} \sum_{i} \sum_{j \neq i} k(\mathbf{v_{i}}, \mathbf{v_{j}})}
\end{split}
\end{equation}

We use a linear kernel for MMD calculations in Eq~\ref{eqn:mmd_final}. The $MMD^2$ measure (MMD for simplicity) can take a big range of values with linear kernel. We report normalized values in range [0,1]. The MMD-test is only carried out on features. The labels are excluded.

\subsubsection{Global-Local context similarity}

\begin{figure*}[!htb]%
    \centering
    \centerline{\includegraphics[width=1\textwidth]{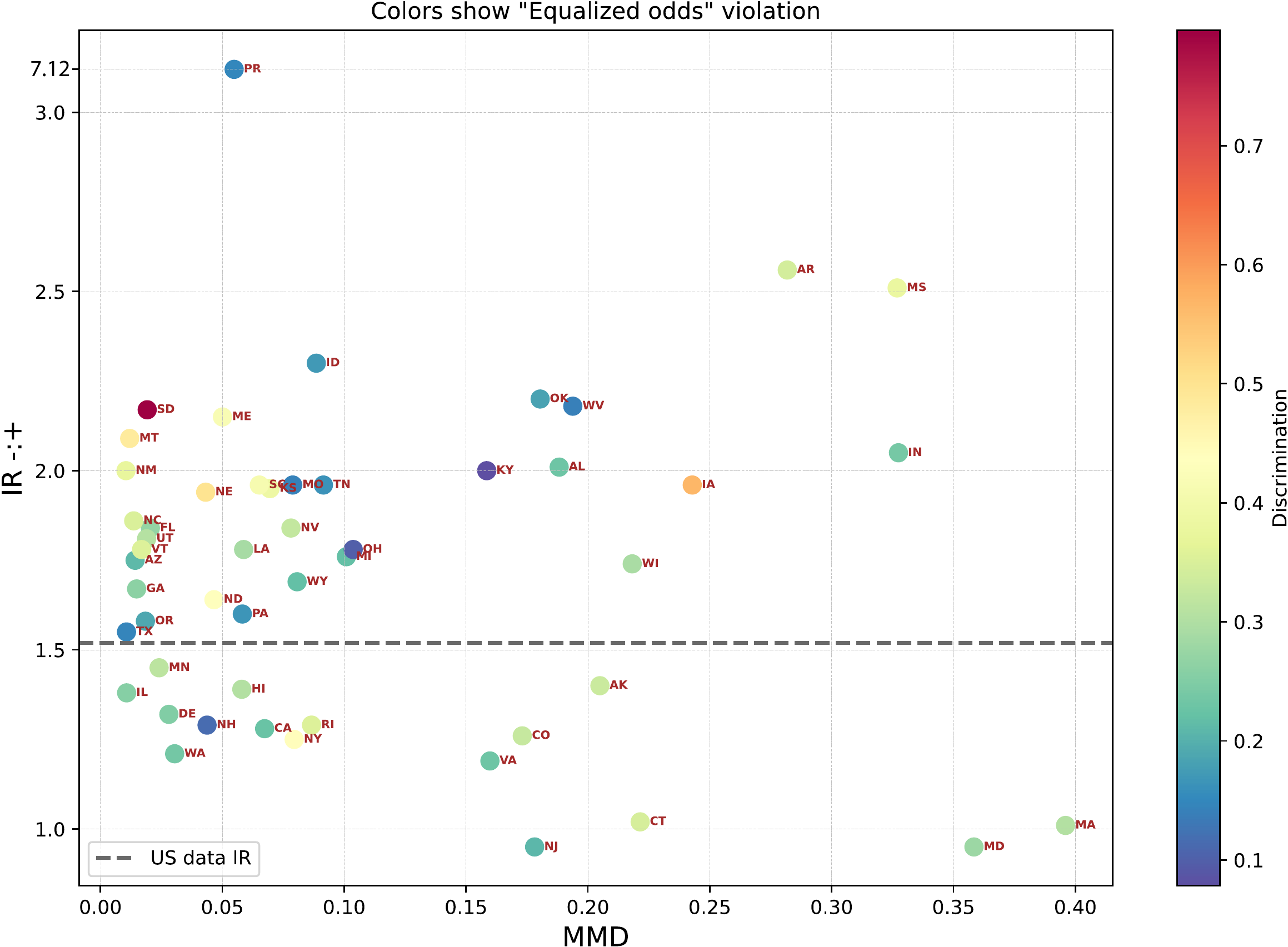}}%
    \caption{Context similarity (MMD) between the global/ ``the US'' dataset and local/context-specific datasets. The IR values (y-axis) refer to the local datasets, the global IR is shown by the dotted line. The z-axis shows the discrimination of the global model deployed on local data (also indicated in color)}
    \label{fig: mmd}
    \vspace{-4mm}
\end{figure*}

Taking the global data as reference, the results of MMD test comparing the global data to each state is shown in Figure~\ref{fig: mmd}. This figure shows the feature-label distribution of datasets comparing to the global data. The horizontal axis indicates the MMD-statistic values and the vertical axis reports Imbalance Ratio (short IR) of +:- and colors indicate ``Eq.Odds'' of global model deployed on each local data. Let us look at VA state in the bottom center toward left and compare it to IN in the right corner. VA has much better MMD-label scores than IN. Now refer back to $\delta$FPR green boxplot in the Figure~\ref{Fig: bp_n2}. VA performs better because it has more similarity to global data. The same applies to the green boxplot in $\delta$FNR and the two pink boxplots. Namely, comparing WA and FL to MT, SD, and IN; the first two dominate the others either in MMD, or in label ratio or in both. 

Although Figure~\ref{fig: mmd} corresponds to a global model (deployed on local context), but it also describes the highlighted states in the local models boxplots (in Figure~\ref{Fig: bp_n2}) well. For example starting from blue boxplots generalizing to red ones, states DE, CA, HI that report lower scores are located in bottom left of context similarity chart but AL, AK, WV, CT, NJ are located farther away with lower similarity to global data. Note that, as discussed before PR (on top left corner in the similarity chart) has a very high IR and its low $\delta$FNRs are due to missclassiying almost all positive instances (from both protected and non-protected groups).

The pattern between context similarity and discrimination scores also exists for the Fairness-aware model results in Figure~\ref{Fig: fair_bp_n2}. In particular comparing FL to SD (pink boxplots), and also for NY, WA compared to AR, SD, IA, IN, MT, and MS (green boxplots). The majority of states are distributed along four bottom-left grids with $MMD<0.10$ and $label-ratio<2$. The ones with lower Eq.Odds violation scores mainly lie below $label-ratio<1.8$ (24 out of 51). Hence, these four grids inhabit the best results (best MMDs, best IRs and apparantely good Eq.Odds scores).

\subsubsection{Local-Local similarity}
\begin{figure*}[!htb]%
    \centering
    \centerline{\includegraphics[width=1\textwidth]{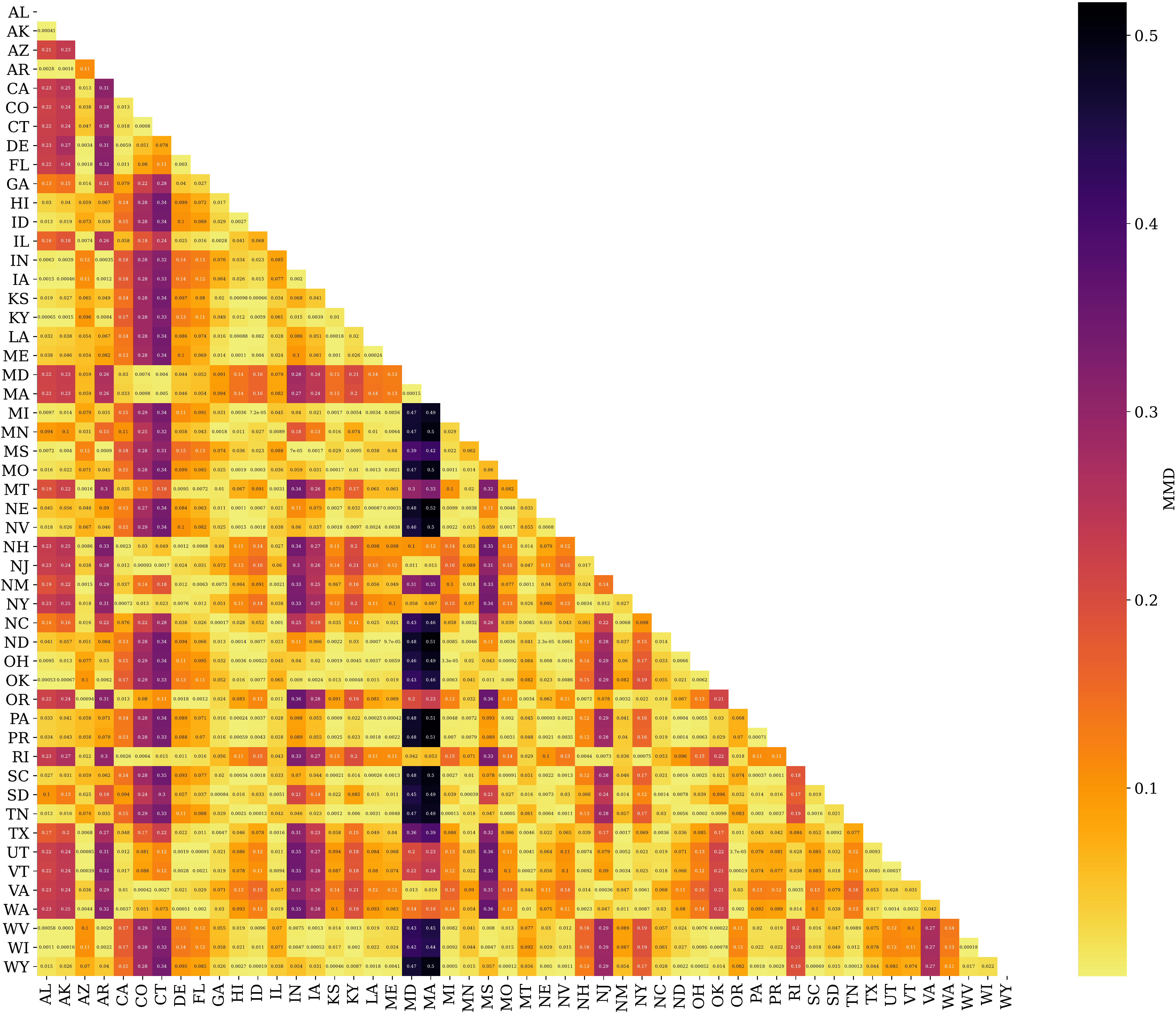}}%
    \caption{Pairwise context similarity (MMD) between states. Light colors indicate higher similarity, dark stand for smaller similarity.}
    \label{fig: mmd_local}
    \vspace{-4mm}
\end{figure*}

To have better insight into distribution of each context, we provide pairwise MMD-test results comparing each pair of contexts/states. Figure~\ref{fig: mmd_local} summarizes the the results. Light colors indicate higher similarity and contrarily darker colors for less similarity. MA, MD seem to have very high dissimilarity rates. As a measure of comparability we sum up the pairwise MMD-score of each state compared to all the other states. This way we can summarize Figure~\ref{fig: mmd_local} and also the states with lowest and highest overall statistical distance from other states. The results are shown in Table~\ref{table:mmd}. From the table we see that NC has the lowest overall distance to the other local contexts, MA has the highest distance and ID is in the middle. If we look back at global-local MMD-Test in Figure~\ref{fig: mmd} these states are also in the similar situation in terms of distance from the global context. MA, MD are located in the bottom-right corner farthest states from the global data (highest MMDs also to global data). NC is located in one of the four best grids and ID is in a medium coordinate. 

\begin{table}%[!htb]%
    \small
    \centering
    \setlength{\tabcolsep}{8pt}
    \caption {\label{table:mmd}Summary of MMD results in Figure~\ref{fig: mmd_local}} 
\begin{tabularx}{0.8\linewidth} {lccc}
%{@{}l*{10}{l}l@{}}
    \toprule\toprule
    &
    \bf{State} &
    \bf{MMD sum} &
    \bf{Eq.Odds} 
    \\
    \midrule

    Best (minimum MMD)  & NC & 1.81 & 0.3514\\
    Worst (maximum MMD) & MA & 17.72 & 0.3048\\
    Median MMD state    & ID & 3.00 & 0.1725\\
    
    \bottomrule\bottomrule
\end{tabularx}
\vspace{-4mm}
\end{table}

\textit{Summary:}, in this section we addressed (\textbf{RQ3}). We first analyzed the distribution similarity/distance of each local dataset to the global data in terms of MMD-statistic and IR. This comparison described very well the behavior of global model deployed on local data in Section~\ref{exp:LocalGlobalModels} and even the behavior of local models to some extent. Moreover, in a second experiment, we studied the pairwise-similrity of local datasets from eachother and realized that, there is a high correlation between global-local MMD-results and the local-local results.

%% file: Conclusion.tex
In this paper, we performed a case-study on the new Adult datasets and conducted a series of experiments to highlight the importance of (spatial) context for fairness to show how distribution shifts affect a model’s performance. Through a variety of experiments we measured fairness performance of LR-models and compared their generalazibility to out-of-distribution deployment sets. Moreover, we studied the impact of fairness interventions on locally vs globally trained models and highlighted the best/worst resulting models and described the reason of good/bad performance based on data and label distribution similarities. We showed that model performance is relative to the degree of similarity between the training and deployment context using an information theoretic two-sample statistical test. However, our work has also some limitations: We experimented based on a limited number of group fairness measures while different measures address different aspects of results and the challenges that data has. In addition, we only studied the behavior of LR classifiers, although the results might slightly differ for other methods but we believe that the final pattern in similarity of results would remain the same. As future directions, we encourage to extend our analysis to temporal aspects of the datasets as well. Another interesting idea is to use the MMD (also other relevant) context similairty metric as a score to build a cluster of augmented local datasets with the final goal t outperform the global model.

%% file: NoBIAS.tex
This work has received funding from the European Union’s Horizon 2020 research and innovation programme under Marie Sklodowska-Curie Actions (grant agreement number 860630) for the project ‘’NoBIAS - Artificial Intelligence without Bias’’.This work reflects only the authors’ views and the European Research Executive Agency (REA) is not responsible for any use that may be made of the information it contains.

%The work of S. Ghodsi, H. Alani, and E. Ntoutsi \todo{Siamka it is only you funded.} has received funding from the European Union’s Horizon 2020 research and innovation programme under Marie Sklodowska-Curie Actions (grant agreement number 860630) for the project “NoBIAS - Artificial Intelligence without Bias”. This work reflects only the authors’ views and the European Research Executive Agency (REA) is not responsible for any use that may be made of the information it contains.